\documentclass[11pt]{article}

\usepackage[final]{acl}

\usepackage{times}
\usepackage{latexsym}
\usepackage[T1]{fontenc}
\usepackage[utf8]{inputenc}
\usepackage{microtype}
\usepackage{inconsolata}
\usepackage{graphicx}
\usepackage{booktabs}
\usepackage{amsmath}
\usepackage{amssymb}
\usepackage{xcolor}
\usepackage{multirow}

\graphicspath{{figures/}}

\newcommand{\method}{\textsc{JustMem}}
\newcommand{\lookup}{\textsc{Lookup}}
\newcommand{\compose}{\textsc{Compose}}
\newcommand{\replay}{\textsc{Replay}}

\title{JustMem: Just-Enough Memory Access for Long-Term Conversations}

\author{
Guanhua Chen \quad
Yanting Wang \quad
Wenjing Zhi \quad
Lei Sha\thanks{Corresponding author.} \\
Beihang University
}

\begin{document}
\maketitle

% ============================================================
% Abstract
% ============================================================

\begin{abstract}
Efficient long-term conversational memory requires retrieving sufficient
evidence without indiscriminately expanding the context presented to
the language model.
This is challenging because relevant evidence may be distributed
across multiple sessions, while compression may discard details needed
for answering.
Different queries therefore require different forms of memory access.
To capture these demands, we formulate memory access along
two dimensions: \emph{discovery breadth}, which controls how broadly
evidence is searched, and \emph{reading fidelity}, which controls
whether evidence is read in compact form or recovered from the
original conversation.
Based on this formulation, we introduce \method{}, which stores
conversation history as compact atomic memories and adapts memory
access along these two dimensions to each query.
Specifically, \lookup{} handles local evidence,
\compose{} broadens discovery for distributed evidence,
and \replay{} increases reading fidelity for fidelity-sensitive evidence.
On LoCoMo and LongMemEval-S, \method{} achieves the highest mean
accuracy and retrieval recall among the compared memory systems while
using substantially fewer generative-model tokens for memory
construction and inference.
\end{abstract}

\section{Introduction}

\begin{figure}[t!]
\centering
\includegraphics[width=\columnwidth]{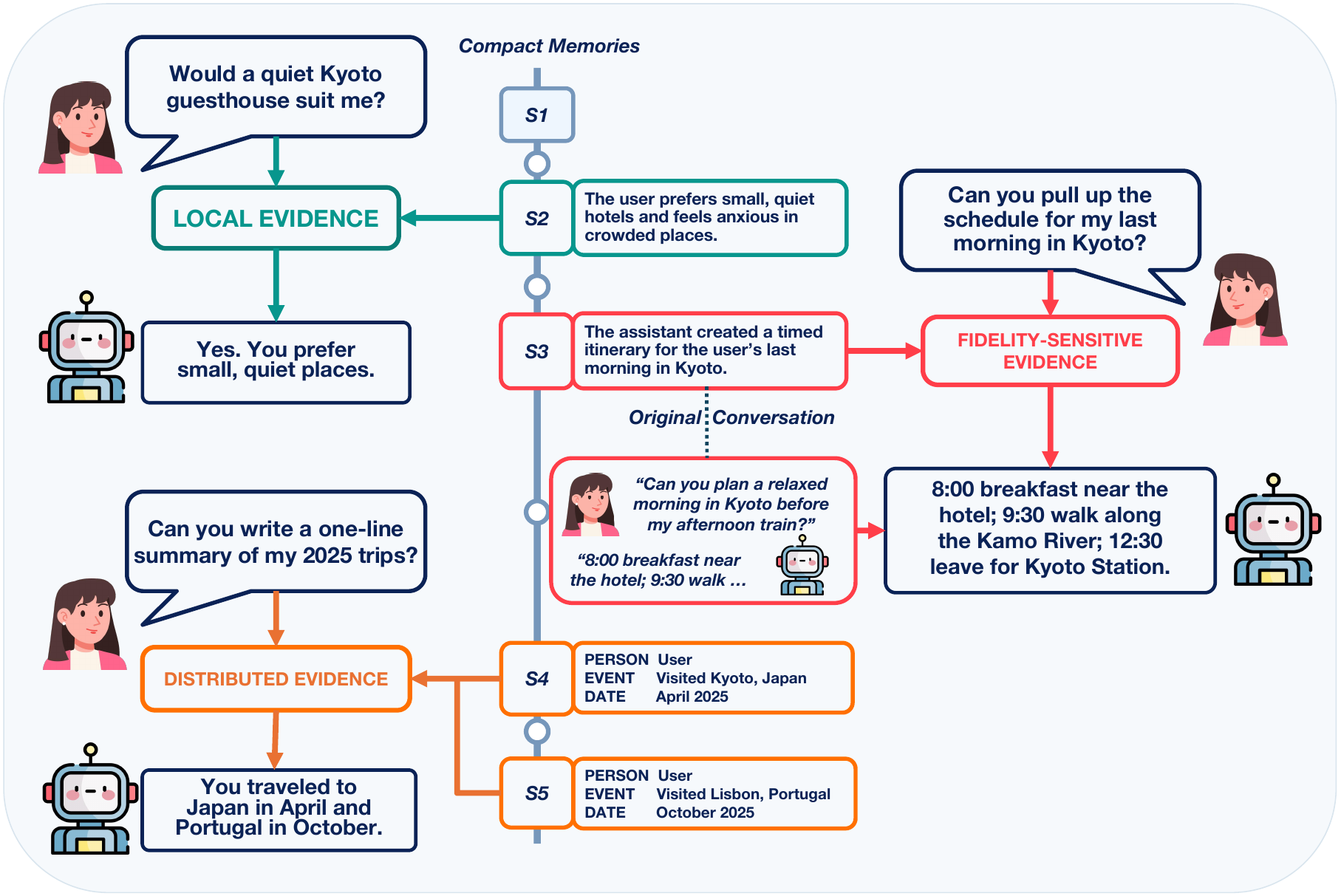}
\caption{
Different questions over the same conversation history impose
different evidence requirements.
Local evidence can be obtained from compact memory, distributed
evidence spans multiple memories, and fidelity-sensitive evidence
may require details from the original conversation.
}
\label{fig:leadin}
\end{figure}

Language-model assistants increasingly operate across conversations
that extend beyond a single context window.
Effective long-term memory must recover prior information, integrate
evidence across sessions, and retain details from earlier interactions.
As illustrated in Figure~\ref{fig:leadin}, these needs correspond to
different evidence requirements.
Some questions depend on \emph{local evidence} captured by compact
memory; others require \emph{distributed evidence} scattered across
multiple sessions; and some require \emph{fidelity-sensitive evidence}
whose answer-critical details may be lost during compression.

Existing memory systems explore persistent stores, summarization,
retrieval augmentation, and explicit memory management
\citep{lewis2020retrieval,zhong2023memorybank,packer2023memgpt}.
However, a largely fixed access policy cannot accommodate these
requirements efficiently.
Narrow retrieval may miss distributed evidence, while compact
representations may omit details needed for high-fidelity recall.
Uniformly expanding retrieval or source context can address these
limitations, but increases context cost and distractor exposure.
The challenge is therefore to provide sufficient evidence without
applying the same access strategy to every query.

We address this challenge by separating memory access along two
dimensions: \emph{discovery breadth}, which controls how broadly
candidate evidence is considered, and \emph{reading fidelity}, which
controls whether compact memory is sufficient or the original
conversation should be consulted.
This separation allows the system to search more broadly without
increasing the final compact evidence budget, while recovering
original content only when compression has removed necessary detail.

Building on this formulation, \method{} provides three
query-conditioned access configurations: \lookup{} for local evidence,
\compose{} for distributed evidence, and \replay{} for
fidelity-sensitive evidence.
All three operate over the same compact atomic memory store, with
broader discovery or source recovery invoked according to query demand.

Experiments on LoCoMo
\citep{maharana-etal-2024-evaluating}, LongMemEval-S
\citep{wu2025longmemeval}, and LoCoMo-Plus
\citep{li-etal-2026-locomo} show that \method{} achieves strong
performance with substantially smaller generative budgets than
existing memory systems.
Controlled experiments further isolate the benefits of global atomic
access, query-guided discovery breadth, and adaptive source recovery.

Our main contributions are:
\begin{itemize}
    \item We formulate long-term conversational memory as adaptive
    evidence allocation along two dimensions: discovery breadth and
    reading fidelity.

    \item We introduce \method{}, a recoverable atomic-memory
    architecture that adapts memory access to local, distributed, and
    fidelity-sensitive evidence demands.

    \item We provide controlled evaluations of memory granularity,
    discovery breadth, and reading fidelity, showing that adaptive
    access achieves strong performance while substantially reducing
    generative budgets.
\end{itemize}

% ============================================================
% Related Work
% ============================================================

\section{Related Work}

\paragraph{Memory representation and maintenance.}
Persistent memory has been implemented through external memory banks,
salience and forgetting mechanisms, and explicit movement between
memory tiers
\citep{zhong2023memorybank,packer2023memgpt}.
Recent systems organize memory as linked notes, hierarchical
short-to-long-term stores, or evolving event structures
\citep{xu2025amem,kang-etal-2025-memory,banerjee-etal-2026-apex}.
Generative and reflective agents also consolidate experience into
higher-level memories
\citep{park2023generative,tan-etal-2025-prospect}.

These approaches primarily improve how conversational experience is
represented, organized, or maintained over time.
\method{} instead focuses on query-time access: it uses a compact,
recoverable memory representation and adapts how that representation
is accessed to the evidence demand of each query.

\paragraph{Memory retrieval and reading.}
RAG conditions generation on retrieved external evidence
\citep{lewis2020retrieval}, while long-context studies show that
adding relevant text does not guarantee reliable use when evidence
competes with distractors \citep{liu2024lost}.
Recent memory systems introduce adaptive graph traversal,
event-to-turn hierarchies, multi-tool retrieval, or coarse retrieval
followed by semantic reranking
\citep{van-etal-2026-memorai,cao-etal-2026-higmem,
banerjee-etal-2026-apex,zhang-etal-2026-lightweight}.

These approaches motivate flexible memory access, but broader
retrieval and higher-fidelity reading impose different costs and
address different evidence failures.
\method{} distinguishes these demands through discovery breadth and
reading fidelity: broad discovery allows more evidence to compete
before selection under a fixed compact read budget, while selective
source recovery introduces higher-fidelity context only when needed.
This allocation provides a path to stronger memory access without
uniformly increasing generative context.
% ============================================================
% JustMem
% ============================================================

\section{JustMem}

Long-term conversational queries differ not only in \emph{what}
evidence they require, but also in \emph{how} that evidence should be
accessed.
Some questions depend on localized evidence, others require evidence
distributed across memories or sessions, and some require details that
compact memory may no longer preserve, such as previously generated
plans, code, or structured assistant outputs.

We characterize these demands along two dimensions:
\emph{discovery breadth}, which determines how broadly candidate
evidence is searched, and \emph{reading fidelity}, which determines
whether compact memory is sufficient or source-level conversational
content is required.
Broader discovery is useful when relevant evidence is dispersed across
memory, whereas higher-fidelity reading is needed when answer-critical
detail may be lost during compression.

\begin{figure}[t]
\centering
\includegraphics[width=\columnwidth]{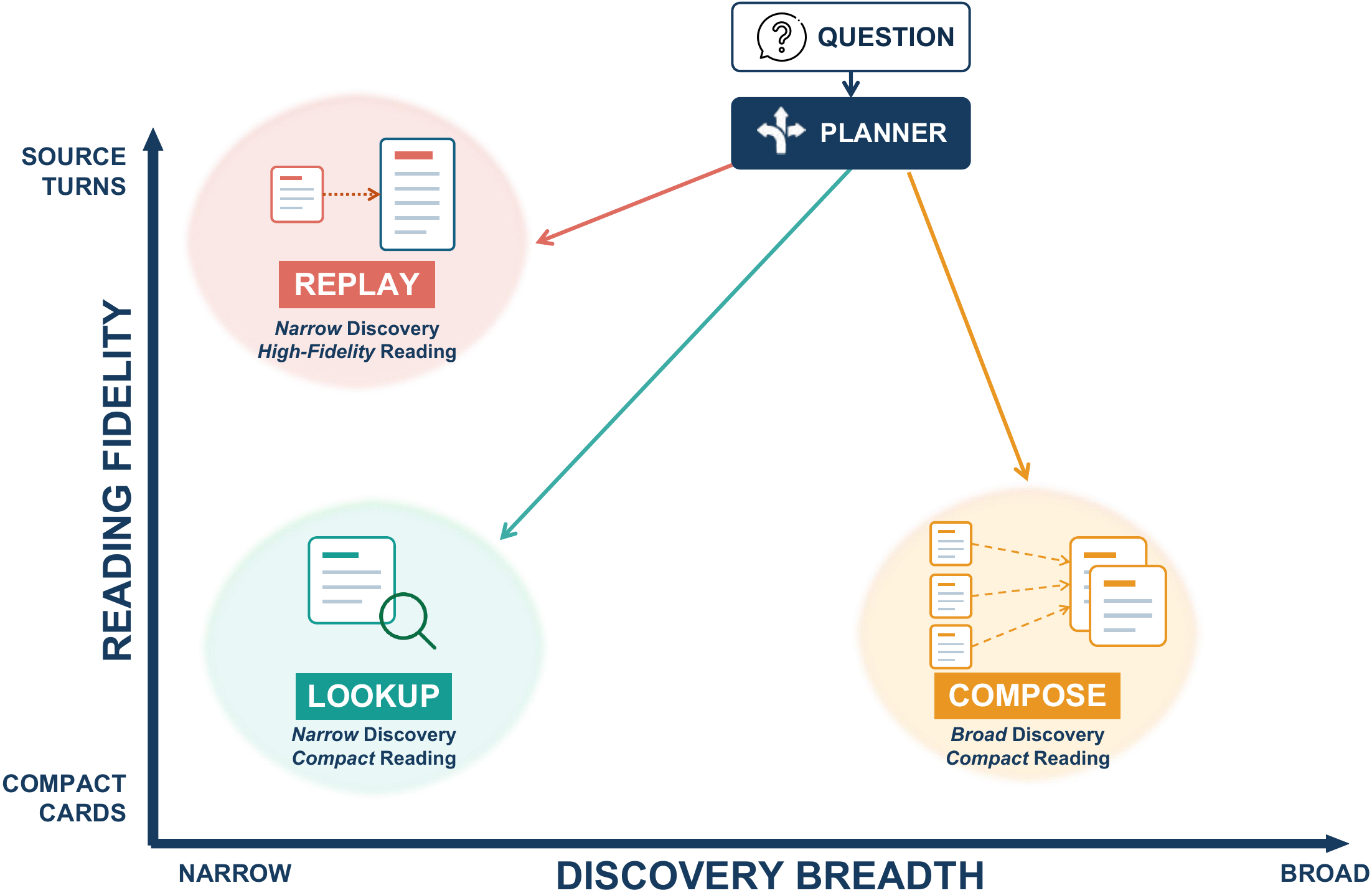}
\caption{
Two dimensions of memory access in \method{}.
\lookup{} uses localized discovery and compact reading;
\compose{} broadens discovery while retaining a bounded compact
evidence budget; and
\replay{} increases reading fidelity through selective source
recovery.
}
\label{fig:allocation-plane}
\end{figure}

Figure~\ref{fig:allocation-plane} shows how \method{} instantiates
these dimensions through three query-conditioned access configurations.
\lookup{} handles \emph{local evidence} with localized discovery and
compact reading.
For \emph{distributed evidence}, \compose{} broadens candidate
discovery while retaining a bounded compact evidence budget.
For \emph{fidelity-sensitive evidence}, \replay{} uses compact
memories to locate provenance-linked source content for higher-fidelity
reading.
The central principle is just-enough memory access: broaden discovery
or recover source content only when required by the query.

\begin{figure*}[t]
\centering
\includegraphics[width=0.82\textwidth]{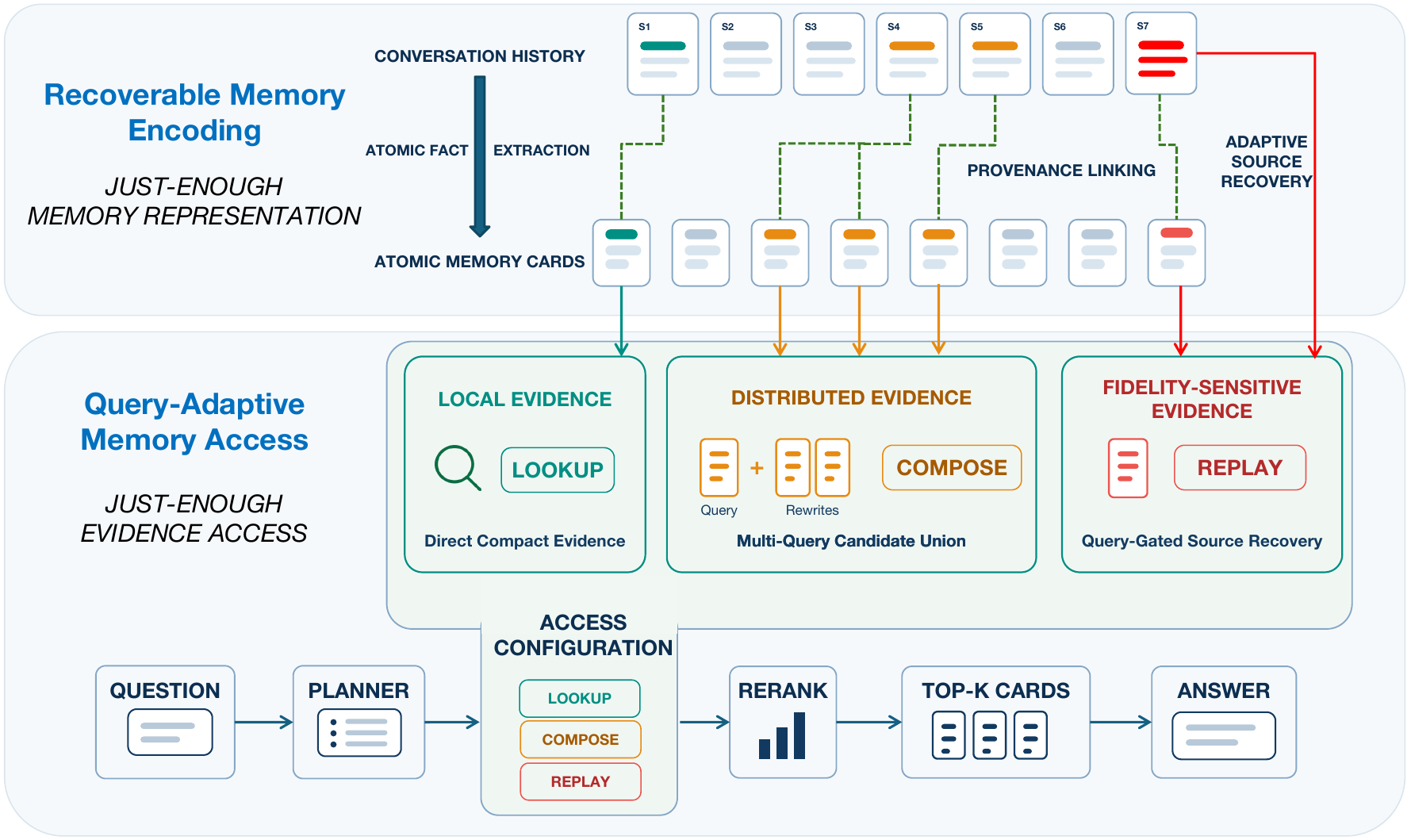}
\caption{
Overview of \method{}.
Conversation history is encoded as compact atomic memories with
provenance links to their source conversations.
At inference time, a structured planner estimates the evidence demand
and selects an access configuration:
\lookup{} for local evidence,
\compose{} for distributed evidence, and
\replay{} for fidelity-sensitive evidence.
The selected evidence is provided to the answer model under a bounded
compact-memory budget, with source content recovered when
higher-fidelity reading is required.
}
\label{fig:method-overview}
\end{figure*}

Building on this allocation view, \method{} organizes memory access
around a compact but recoverable memory store, as shown in
Figure~\ref{fig:method-overview}.
At query time, it predicts the evidence requirements and selects how
that memory should be accessed.
All three configurations operate over the same atomic memory
representation.

% ------------------------------------------------------------
% Recoverable Memory Encoding
% ------------------------------------------------------------

\subsection{Recoverable Memory Encoding}

For each conversation session, \method{} extracts a set of atomic
memory cards from the conversational contents.
Each card contains compact memory content, temporal information, a
provenance link to its source conversation, and lightweight metadata
describing its memory type and access attributes.
All cards and their attributes for a session are produced jointly in
a single extraction-model call.

A session may yield multiple cards corresponding to distinct facts,
events, preferences, states, or interactions.
We use \emph{atomic} in the retrieval sense: each card represents a
piece of information that can be retrieved and interpreted
independently.
This granularity allows individual memories, including multiple cards
from the same session, to compete globally during retrieval.

Memory types retain the structure needed for subsequent access.
Factual memories are stored as subject-explicit natural-language
statements, while event-oriented memories additionally preserve
explicit subject and temporal information for compositional access.

Provenance makes the compact representation recoverable.
A card can serve directly as evidence when its content is sufficient,
while its provenance provides access to the original conversation when
source-level detail is required.
Section~\ref{sec:granularity-ablation} isolates the effect of atomic
retrieval granularity while holding memory content fixed.

% ------------------------------------------------------------
% Query-Adaptive Memory Access
% ------------------------------------------------------------

\subsection{Query-Adaptive Memory Access}

Given a query, \method{} first predicts its evidence requirements,
executes one of three access configurations, and provides the selected
evidence to the answer model.
The answer is grounded in this evidence, with multiple memories
combined according to the planned operation when needed.

\subsubsection{Access Planning}

The planner produces a structured retrieval specification containing
target entities, temporal constraints, and the operation to be
performed over retrieved evidence.
These fields specify the relevant evidence scope and how selected
evidence should be used, including direct lookup, aggregation,
temporal selection, ordering, or comparison.

The planner additionally predicts two binary access demands, which
determine the access configuration:
\begin{equation}
z_q =
\begin{cases}
\replay,  & f_q = 1,\\
\compose, & f_q = 0 \land d_q = 1,\\
\lookup,  & \text{otherwise}
\end{cases}
\label{eq:access-routing}
\end{equation}
where $z_q$ is the access configuration, and $d_q,f_q\in\{0,1\}$
indicate distributed-evidence and source-fidelity demand, respectively.

For distributed-evidence queries, the planner also generates up to two
answer-free retrieval rewrites that provide complementary views of the
same information need while preserving the entities and constraints
of the original query.
The retrieval specification, access-demand indicators, and applicable
rewrites are produced jointly in a single planner call.
The access configuration is then assigned by the deterministic rule
in Eq.~\ref{eq:access-routing}, without an additional model call.

Thus, the model predicts structured evidence requirements, while a
fixed rule maps those requirements to an interpretable access
configuration.
Section~\ref{sec:access-allocation} compares this adaptive allocation
with fixed \lookup{}, \compose{}, and \replay{} configurations.

\subsubsection{\lookup{}: Local Evidence}

For queries supported by localized evidence, \lookup{} retrieves
candidate memories using semantic similarity to the original query.
The candidates are reranked against the query and reduced to a bounded
set of compact memory cards.

\lookup{} provides the lightweight access path when answer-bearing
evidence can be localized and represented adequately by compact
memory.

\subsubsection{\compose{}: Distributed Evidence}

For queries whose evidence may be distributed across memories or
sessions, \compose{} broadens candidate discovery before selection.
It forms a multi-view candidate union:
\begin{equation}
C_q =
C(q)\cup\bigcup_{r\in R_q} C(r),
\label{eq:compose}
\end{equation}
where $C(x)$ is the candidate set for retrieval view $x$, $R_q$ is the
rewrite set for query $q$, and $C_q$ is their deduplicated union.
Operation-aware routing determines the relevant candidate scope, while
the rewrites provide complementary retrieval views.

The resulting candidates are reduced to a fixed compact evidence
budget:
\begin{equation}
E_q =
\operatorname{Top}_{K}
\left(
\operatorname{Rerank}(q,C_q)
\right),
\label{eq:compact-read}
\end{equation}
where $\operatorname{Rerank}(q,C_q)$ orders candidates against $q$,
$\operatorname{Top}_{K}$ retains the best $K$ cards, and $E_q$ is the
resulting compact evidence set.
Broader discovery therefore changes which memories compete for
selection without increasing the number of compact cards presented to
the answer model.
Section~\ref{sec:breadth-ablation} evaluates this effect while holding
the reranker and final evidence budget fixed.

\subsubsection{\replay{}: Fidelity-Sensitive Evidence}

Some queries require information that compact memory can locate but
cannot preserve faithfully.
Typical examples include previously generated code, detailed plans,
or structured assistant outputs whose answer-critical content is not
fully retained in a short memory card.

For these queries, \replay{} retrieves and reranks compact memories to
identify relevant source conversations.
The selected cards serve both as compact evidence and as provenance
anchors for source recovery.

Reading fidelity is controlled at the query level.
When the planner predicts a fidelity-sensitive demand ($f_q=1$),
\replay{} follows the provenance of the selected memories to recover
source-level conversational content; otherwise, the answer model reads
only the compact evidence.
Section~\ref{sec:fidelity-ablation} compares this query-gated policy
with fixed compact reading and fixed source recovery.
% ============================================================
% Experimental Setup
% ============================================================
\begin{table*}[t]
\centering
\small
\setlength{\tabcolsep}{8pt}
\begin{tabular}{@{}cccccc@{}}
\toprule
Dataset & Method & Recall@10 (\%) $\uparrow$ & Acc. (\%) $\uparrow$
& Construction (K) $\downarrow$ & Infer (K) $\downarrow$ \\
\midrule
\multirow{7}{*}{LoCoMo}
& FullText
    & -- & $57.18 \pm 0.09$ & -- & 16.74 \\
& Naive RAG
    & 57.91 & $63.82 \pm 0.10$ & -- & 0.63 \\
\cmidrule(lr){2-6}
& A-Mem
    & 79.74 & $67.91 \pm 0.27$ & 1188.64 & 21.87 \\
& MemoryOS
    & 41.85 & $65.12 \pm 0.26$ & 382.76 & 5.11 \\
& SimpleMem
    & 75.83 & $76.55 \pm 0.24$ & 96.17 & 5.42 \\
& LightMem
    & 81.92 & $79.08 \pm 0.26$ & 108.21 & 4.06 \\
& \textbf{JustMem}
    & \textbf{88.05}
    & $\mathbf{79.61 \pm 0.12}$
    & \textbf{53.38}
    & \textbf{1.33} \\
\midrule
\multirow{7}{*}{LongMemEval-S}
& FullText
    & -- & $77.16 \pm 0.09$ & -- & 111.84 \\
& Naive RAG
    & 80.74 & $67.88 \pm 0.11$ & -- & 12.08 \\
\cmidrule(lr){2-6}
& A-Mem
    & 92.47 & $71.68 \pm 0.30$ & 1321.42 & 15.63 \\
& MemoryOS
    & 81.12 & $67.56 \pm 0.36$ & 674.35 & 9.07 \\
& SimpleMem
    & 94.83 & $74.92 \pm 0.39$ & 159.72 & 9.28 \\
& LightMem
    & 89.91 & $76.44 \pm 0.17$ & 152.08 & 3.84 \\
& \textbf{JustMem}
    & \textbf{96.00}
    & $\mathbf{83.40 \pm 0.24}$
    & \textbf{53.33}
    & \textbf{1.70} \\
\bottomrule
\end{tabular}
\caption{
Matched-configuration comparison on LoCoMo and LongMemEval-S.
Accuracy is reported as mean $\pm$ sample standard deviation over five
answering-and-evaluation runs, with constructed memories and retrieved
evidence held fixed.
Recall@10 measures retrieval recall over the Top-10 selected evidence.
Construction and Infer denote generative-model token usage for memory
construction and online query processing, respectively, reported in
thousands.
Recall@10 is not applicable to FullText because it does not perform
Top-10 retrieval.
FullText and Naive RAG are reference systems and are excluded from
best-value highlighting.
}
\label{tab:main}
\end{table*}
\section{Experimental Setup}
\label{sec:experimental-setup}

\subsection{Datasets}

We evaluate \method{} primarily on two long-term conversational memory
benchmarks.
LoCoMo \citep{maharana-etal-2024-evaluating} evaluates memory over
long multi-session conversations, including factual recall, temporal
reasoning, and cross-session information integration.
We use the 1,540 questions from Categories 1--4; Category 5 is
excluded because it does not provide the gold-answer annotations
required by our answer-accuracy evaluation.

LongMemEval-S \citep{wu2025longmemeval} covers complementary memory
capabilities, including knowledge updates, multi-session reasoning,
temporal reasoning, user preferences, and recall of prior user and
assistant content.
We use all 500 questions from LongMemEval-S.

To further evaluate cross-benchmark transfer, we use the 401-question
cognitive split of LoCoMo-Plus \citep{li-etal-2026-locomo}.

\subsection{Evaluation Protocol}

We report LLM-judge accuracy as the primary metric and Recall@10 for
retrieval quality, defined as the fraction of gold evidence recovered
within the Top-10 selected evidence.
Paired comparisons use wrong-to-right and right-to-wrong flips with
the exact McNemar test \citep{mcnemar1947note}.

For all methods, we repeat answering and judging five times while
keeping constructed memories and retrieved evidence fixed.
Table~\ref{tab:main} reports mean accuracy $\pm$ sample standard
deviation across these runs.

We measure efficiency by generative-model token usage.
Construction counts input and output tokens used by the memory-construction model for a complete conversation history, while Infer
counts query planning and answer generation per question; embedding,
reranking, and offline evaluation are excluded.
Reading-fidelity experiments additionally report answer-stage tokens
to capture the context introduced by source recovery.

\subsection{Implementation and Comparison Settings}

\method{} uses GPT-4.1-mini \citep{openai2025gpt41} for memory
extraction, query planning, and answer generation.
Memory cards are indexed using 1,536-dimensional
text-embedding-3-small embeddings \citep{openai2024embeddings}, and
retrieved candidates are reranked with Qwen3-Rerank
\citep{zhang2025qwen3embedding}.
Unless otherwise specified, all retrieval-based methods use $K=10$
as the final evidence budget.
For \method{}, this corresponds to ten compact memory cards.
The budget remains fixed when discovery is broadened; additional
answer-stage context is introduced only through source recovery.

We compare against two reference settings and four representative
long-term memory systems.
FullText provides the complete conversation history to the answer
model, while Naive RAG retrieves conversation content using embedding
similarity.
A-Mem \citep{xu2025amem} organizes memories as dynamically linked
notes, MemoryOS \citep{kang-etal-2025-memory} uses hierarchical memory
tiers, SimpleMem \citep{liu2026simplemem} constructs compact semantic
memories, and LightMem \citep{zhang-etal-2026-lightweight} combines
lightweight memory construction with retrieval and reranking.
Together, these baselines cover full-context access, standard
retrieval augmentation, and recent structured or compressed memory
architectures.

For LoCoMo and LongMemEval-S, all methods are evaluated under the same
model configuration.
GPT-4.1-mini is used consistently for generative memory construction,
answer generation, and LLM-based evaluation across all methods, while
text-embedding-3-small is used for embedding-based retrieval.
All methods use the same dataset-specific evaluation prompts.
This matched setting reduces variation from memory-construction,
answer, evaluation, and embedding models, allowing the comparison to
focus on differences in memory representation and access strategy.

For LoCoMo-Plus, all methods follow the original benchmark protocol
\citep{li-etal-2026-locomo}, with GPT-4.1-mini used for generative
memory construction, answer generation, and evaluation to match our
primary experiments.

% ============================================================
% Results
% ============================================================

\section{Results}

We evaluate \method{} end to end and test the three hypotheses
underlying its design: atomic access granularity (H1), discovery
breadth under a fixed compact evidence budget (H2), and query-gated
reading fidelity (H3).
We then examine adaptive access allocation and cross-benchmark
transfer to LoCoMo-Plus.

% ============================================================
% Overall Performance
% ============================================================

\subsection{Overall Performance}
\label{sec:overall-results}

Table~\ref{tab:main} compares \method{} with existing memory systems
under the matched configuration described in
Section~\ref{sec:experimental-setup}.
\method{} achieves the highest mean accuracy and Recall@10 on both
LoCoMo and LongMemEval-S, with small variation across repeated
answering-and-evaluation runs.
These gains are achieved with the smallest generative budgets among
the compared memory systems.

Relative to LightMem, the strongest competing memory system by mean
accuracy (paired $p=0.532$ on LoCoMo), \method{} reduces
memory-construction and query-time token usage by 50.7\% and 67.2\%
on LoCoMo, and by 64.9\% and 55.7\% on LongMemEval-S, respectively.
The results therefore show that stronger memory performance does not
require larger generative budgets for either memory construction or
query-time inference.

% ============================================================
% H1: Atomic Granularity
% ============================================================

\subsection{H1: Atomic Granularity Enables Global Access}
\label{sec:granularity-ablation}

We test whether atomic memories are most effective when they can
compete directly across the full memory store.
Table~\ref{tab:atomicity} compares three access strategies constructed
from the same 3,804 atomic cards.
\textbf{Session pack Top-1} concatenates all cards from each session
into a single retrieval unit and provides the highest-ranked session
to the answer model.
\textbf{Session-first atom Top-10} first ranks sessions and selects the
highest-ranked atomic candidate from each session, enforcing
cross-session diversity.
\textbf{Global atom Top-10} directly ranks atomic candidates across
sessions, allowing multiple complementary cards from the same session
to survive selection.

Global atomic access improves accuracy by 10.72 points over
Session pack Top-1 ($p=2.65\times10^{-22}$).
Under the more closely matched Session-first atom Top-10 comparison,
global atomic access improves accuracy by 5.52 points, with 134
wrong-to-right versus 49 right-to-wrong flips
($p=2.54\times10^{-10}$), while using only 1.1\% more answer-stage
tokens.
Because both conditions expose approximately the same amount of atomic
evidence to the answer model, this comparison isolates the benefit of
global atomic competition: relevant evidence can be selected across
session boundaries while multiple complementary memories from the
same useful session can be retained.

% ============================================================
% H2: Discovery Breadth
% ============================================================

\subsection{H2: Query-Guided Breadth Improves Distributed-Evidence Access}
\label{sec:breadth-ablation}

We isolate discovery breadth while holding the atomic memory store,
structured query plan, reranker, answer model, and final Top-10
evidence budget fixed.
As shown in Table~\ref{tab:discovery-breadth}, local discovery
retrieves candidates from the original query, whereas guided expansion
additionally incorporates candidates retrieved from up to two
planner-generated rewrites.
Both candidate pools are reranked against the original question before
the answer model reads exactly ten compact cards.

Guided expansion improves overall accuracy from 75.91\% to 77.86\%
on LoCoMo ($p=0.032$) and from 71.20\% to 76.60\% on LongMemEval-S
($p=3.55\times10^{-4}$).
On LongMemEval-S, Recall@10 also increases from 91.80\% to 96.60\%.
Because the reranker and final reading budget are unchanged, these
results isolate the benefit of allowing evidence from additional
query-guided retrieval views to compete for the same compact context.

The gain is particularly pronounced for aggregation-oriented queries,
which improve from 68.93\% to 77.97\% on LoCoMo and from 65.96\% to
78.72\% on LongMemEval-S.
This pattern supports broader discovery when answering requires
collecting evidence distributed across multiple memories.

% ============================================================
% H3: Reading Fidelity
% ============================================================

\subsection{H3: Reading Fidelity Should Be Query-Gated}
\label{sec:fidelity-ablation}

\begin{table}[t]
\centering
\small
\begin{tabular}{@{}lrr@{}}
\toprule
Access unit & Accuracy & Tok./q \\
\midrule
Session pack Top-1
    & 69.09 & 772.08 \\
Session-first atom Top-10
    & 74.29 & 819.25 \\
Global atom Top-10
    & \textbf{79.81} & 828.28 \\
\bottomrule
\end{tabular}
\caption{
Retrieval-granularity ablation on LoCoMo.
Variants use the same atomic memories but organize and select them
differently for retrieval.
Tok./q reports answer-stage token usage.
}
\label{tab:atomicity}
\end{table}

\begin{table}[t]
\centering
\small
\setlength{\tabcolsep}{3.0pt}
\begin{tabular}{@{}llrr@{}}
\toprule
Dataset & Query group & \shortstack{Local\\Discovery} & \shortstack{Guided\\Expansion} \\
\midrule
\multirow{2}{*}{LoCoMo}
& Overall     & 75.91 & \textbf{77.86} \\
& Aggregation & 68.93 & \textbf{77.97} \\
\midrule
\multirow{2}{*}{LongMemEval-S}
& Overall     & 71.20 & \textbf{76.60} \\
& Aggregation & 65.96 & \textbf{78.72} \\
\bottomrule
\end{tabular}
\caption{
Accuracy (\%) under controlled discovery breadth.
Local Discovery uses the original query, while Guided Expansion
incorporates additional planner-generated retrieval views.
Both settings use the same reranker and final Top-10 evidence budget.
Query groups are derived from the operation predicted during access
planning.
}
\label{tab:discovery-breadth}
\end{table}

\begin{table}[t]
\centering
\small
\setlength{\tabcolsep}{3pt}
\begin{tabular}{@{}lrrr@{}}
\toprule
& \multicolumn{2}{c}{Question Group} & \\
\cmidrule(lr){2-3}
Reading policy & Overall & Asst. & Tok./q \\
\midrule
Compact Reading
    & 74.40 & 32.14 & 910 \\
Fixed Top-1 Recovery
    & 80.60 & 92.86 & 3967 \\
Adaptive \replay{}
    & \textbf{83.40} & \textbf{94.64} & 1155 \\
\bottomrule
\end{tabular}
\caption{
Reading-fidelity ablation on LongMemEval-S.
\textbf{Compact Reading} uses only the selected compact memories;
\textbf{Fixed Top-1 Recovery} restores the highest-ranked
provenance-linked source session for every query; and
\textbf{Adaptive \replay{}} activates source recovery only when
required by the query.
Overall and Asst. report accuracy (\%) on all 500 questions and the
56 assistant-output questions, respectively; Tok./q reports
answer-stage token usage.
}
\label{tab:replay}
\end{table}

We test whether reading fidelity should be fixed across queries or
allocated according to query demand.
Table~\ref{tab:replay} compares three policies:
\textbf{Compact Reading} uses only compact memories,
\textbf{Fixed Top-1 Recovery} restores the highest-ranked
provenance-linked source session for every query, and
\textbf{Adaptive \replay{}} activates source recovery only for
fidelity-sensitive queries.

Source recovery substantially improves fidelity-sensitive recall.
Compared with compact reading, fixed recovery raises overall accuracy
from 74.40\% to 80.60\% ($p=9.26\times10^{-6}$) and assistant-output
accuracy from 32.14\% to 92.86\%, but increases answer-stage context
from 910 to 3967 tokens per question.
Adaptive \replay{} retains this fidelity benefit while restricting
source recovery to queries that require it, reaching 83.40\% overall
accuracy at 1155 tokens per question and significantly outperforming
fixed recovery ($p=0.00258$).

On the assistant-output subset, fixed and adaptive recovery achieve
similar accuracy (92.86\% vs.\ 94.64\%, $p=1.0$).
Thus, the overall advantage of adaptive recovery comes primarily from
avoiding unnecessary source exposure on other queries, rather than
from improving source-based answering itself.
Appendix~\ref{sec:replay-analysis} further shows that once the
answer-bearing source is correctly localized, all such cases are
answered correctly, identifying source localization as the remaining
bottleneck.

% ============================================================
% Controlled Access Allocation
% ============================================================

\subsection{Controlled Access Allocation}
\label{sec:access-allocation}

\begin{table}[t]
\centering
\small
\setlength{\tabcolsep}{3.0pt}
\begin{tabular}{@{}lrrr@{}}
\toprule
Access configuration & Recall@10 & Acc. & Infer (K) \\
\midrule
Fixed \lookup{}
    & 91.80 & 71.20 & 1.47 \\
Fixed \compose{}
    & \textbf{96.60} & 76.60 & 1.69 \\
Fixed \replay{}
    & 87.40 & 63.40 & 5.06 \\
Adaptive access
    & 96.00 & 83.40 & 1.70 \\
Taxonomy oracle$^\dagger$
    & 96.20 & \textbf{84.20} & 1.70 \\
\bottomrule
\end{tabular}
\caption{
Controlled access allocation on all 500 LongMemEval-S questions.
Fixed configurations override access while preserving other planner
outputs. $^\dagger$The taxonomy oracle uses benchmark-provided question
categories, which are privileged at inference time.
}
\label{tab:access-allocation}
\end{table}

We next test whether discovery breadth and reading fidelity should be
allocated by query demand or imposed globally.
Table~\ref{tab:access-allocation} compares five access policies:
\textbf{Fixed \lookup{}}, \textbf{Fixed \compose{}}, and
\textbf{Fixed \replay{}} force every query to use the corresponding
access configuration while preserving the remaining planner outputs;
\textbf{Adaptive Access} uses the query-conditioned routing in
Eq.~\ref{eq:access-routing}; and the \textbf{Taxonomy Oracle} replaces
this routing decision with the benchmark-provided question category.

No fixed configuration performs well across all queries.
Fixed \lookup{} provides the least expensive access but reaches only
71.20\% accuracy, while fixed \compose{} improves evidence coverage to
96.60\% Recall@10 and accuracy to 76.60\%.
Applying \replay{} indiscriminately is substantially less effective:
it reduces Recall@10 to 87.40\%, raises inference cost to 5.06K tokens
per question, and reaches only 63.40\% accuracy.
In contrast, adaptive access reaches 83.40\% accuracy with 96.00\%
Recall@10 at 1.70K inference tokens.

The taxonomy oracle reaches 84.20\% accuracy at the same inference
budget, only 0.8 points above adaptive access.
Because it uses privileged benchmark metadata unavailable at inference
time, it serves only as an oracle reference.
Together with H2 and H3, these results support matching discovery
breadth and reading fidelity to query-specific evidence demands rather
than imposing a single global access policy.

% ============================================================
% Cross-Benchmark Transfer
% ============================================================

\subsection{Cross-Benchmark Transfer to LoCoMo-Plus}
\label{sec:locomo-plus-transfer}

We finally evaluate cross-benchmark transfer on the 401-question
cognitive split of LoCoMo-Plus, which tests whether memory systems can
retain and apply implicit conversational constraints under
cue--trigger semantic disconnect.
Table~\ref{tab:locomo-plus} reports accuracy on both benchmarks
together with the absolute performance drop from LoCoMo to
LoCoMo-Plus.

\method{} achieves 62.59\% accuracy on LoCoMo-Plus, compared with
31.67\% for the strongest baseline, MemOS.
It also exhibits the smallest cross-benchmark gap at 17.02 points,
compared with 33.45--49.71 points for the competing methods.
These results indicate that \method{} retains substantially more of
its performance when moving from conventional long-term memory
questions to the cognitive-memory setting.

This transfer behavior is consistent with the design of \method{}.
LoCoMo-Plus weakens the direct semantic correspondence between a
current query and the earlier conversational cue that should influence
the response, making evidence localization more difficult for fixed
similarity-driven access.
In \method{}, atomic memories provide fine-grained, globally
accessible retrieval units, while query-adaptive access can broaden
candidate discovery beyond the original query when relevant evidence
is difficult to localize.
Provenance-linked source recovery further preserves access to the
underlying conversation when compact memories are insufficient.
The smaller transfer gap therefore suggests that separating evidence
discovery from reading fidelity may provide a more robust access
strategy when query--memory correspondence becomes indirect.

\section{Conclusion}
\begin{table}[t]
\centering
\small
\setlength{\tabcolsep}{5pt}
\begin{tabular}{@{}lrrr@{}}
\toprule
Method & LoCoMo $\uparrow$ & \shortstack{LoCoMo-\\Plus $\uparrow$} & Gap $\downarrow$ \\
\midrule
FullText
    & 57.18 & 21.70 & 35.48 \\
Naive RAG
    & 63.82 & 15.71 & 48.11 \\
\midrule
A-Mem
    & 67.91 & 18.20 & 49.71 \\
MemOS
    & 65.12 & 31.67 & 33.45 \\
\midrule
\textbf{JustMem}
    & \textbf{79.61}
    & \textbf{62.59}
    & \textbf{17.02} \\
\bottomrule
\end{tabular}
\caption{
Accuracy (\%) on LoCoMo and the 401-question cognitive split of
LoCoMo-Plus.
Gap denotes the absolute accuracy drop from LoCoMo to LoCoMo-Plus
(lower is better).
}
\label{tab:locomo-plus}
\end{table}
We formulated long-term conversational memory as adaptive evidence allocation along two dimensions: discovery breadth and reading fidelity.
We instantiated this formulation in \method{}, which combines recoverable atomic memories with query-conditioned reading to search broadly when evidence is distributed and recover source-level detail when compact memory is insufficient.
Across long-term memory benchmarks, \method{} achieves strong performance with substantially smaller generative budgets, while controlled experiments validate the distinct roles of retrieval granularity, discovery breadth, and selective source recovery.
The broader lesson is simple: discover as broadly as the evidence requires, and read only at the fidelity the question demands.

% ============================================================
% Limitations
% ============================================================

\section*{Limitations}

Our evaluation focuses on established long-term conversational memory benchmarks, and broader evaluation across tasks and settings would further assess the generality of JUSTMEM.
The current implementation adopts a specific memory and retrieval configuration, while alternative configurations may offer different effectiveness--efficiency trade-offs.
Finally, our efficiency analysis primarily focuses on generative-model token usage and does not capture all system-level computational costs.

% ============================================================
% Ethical Considerations
% ============================================================

\section*{Ethical Considerations}

Long-term conversational memory introduces privacy, retention, and access-control risks, particularly because \method{} retains provenance links that can recover source conversational content.
Deployed systems should minimize stored information, obtain appropriate user consent, support inspection and deletion, and protect both memory cards and their source content.
Because stored memories may also be incorrect or outdated, selective retrieval and source recovery do not eliminate extraction errors, retrieval bias, or sensitive-information leakage.

\section*{Acknowledgement}

This work was supported by the National Science Fund for Excellent Young Scholars (Overseas) under grant No.\ KZ37117501, National Natural Science Foundation of China ( No. \ 62306024), National Cyber Security-National Science and Technology Major Project (No. 2025ZD1503602), New Generation Artificial Intelligence-National Science and Technology Major Project (No. 2026ZD0128000, No. 2026ZD0128003), the Fundamental Research Funds for the Central Universities,
and Beijing Advanced Innovation Center for Future Blockchain and Privacy Computing.

\bibliography{custom}

@inproceedings{maharana-etal-2024-evaluating,
  title={Evaluating very long-term conversational memory of llm agents},
  author={Maharana, Adyasha and Lee, Dong-Ho and Tulyakov, Sergey and Bansal, Mohit and Barbieri, Francesco and Fang, Yuwei},
  booktitle={Proceedings of the 62nd Annual Meeting of the Association for Computational Linguistics (Volume 1: Long Papers)},
  pages={13851--13870},
  year={2024}
}

@inproceedings{wu2025longmemeval,
  author={Wu, Di and Wang, Hongwei and Yu, Wenhao and Zhang, Yuwei and Chang, Kai-Wei and Yu, Dong},
  title={LongMemEval: Benchmarking Chat Assistants on Long-Term Interactive Memory},
  booktitle={International Conference on Learning Representations},
  editor={Y. Yue and A. Garg and N. Peng and F. Sha and R. Yu},
  pages={86809--86836},
  volume={2025},
  year={2025},
  url={https://proceedings.iclr.cc/paper_files/paper/2025/file/d813d324dbf0598bbdc9c8e79740ed01-Paper-Conference.pdf}
}

@article{packer2023memgpt,
  title={Memgpt: Towards llms as operating systems},
  author={Packer, Charles and Wooders, Sarah and Lin, Kevin and Fang, Vivian and Patil, Shishir G and Stoica, Ion and Gonzalez, Joseph E},
  journal={arXiv preprint arXiv:2310.08560},
  year={2023}
}

@inproceedings{zhong2023memorybank,
  title={Memorybank: Enhancing large language models with long-term memory},
  author={Zhong, Wanjun and Guo, Lianghong and Gao, Qiqi and Ye, He and Wang, Yanlin},
  booktitle={Proceedings of the AAAI conference on artificial intelligence},
  volume={38},
  pages={19724--19731},
  year={2024}
}

@inproceedings{park2023generative,
  title={Generative agents: Interactive simulacra of human behavior},
  author={Park, Joon Sung and O'Brien, Joseph and Cai, Carrie Jun and Morris, Meredith Ringel and Liang, Percy and Bernstein, Michael S},
  booktitle={Proceedings of the 36th annual acm symposium on user interface software and technology},
  pages={1--22},
  year={2023}
}

@article{lewis2020retrieval,
  title={Retrieval-augmented generation for knowledge-intensive nlp tasks},
  author={Lewis, Patrick and Perez, Ethan and Piktus, Aleksandra and Petroni, Fabio and Karpukhin, Vladimir and Goyal, Naman and K{\"u}ttler, Heinrich and Lewis, Mike and Yih, Wen-tau and Rockt{\"a}schel, Tim and others},
  journal={Advances in neural information processing systems},
  volume={33},
  pages={9459--9474},
  year={2020}
}

@article{liu2024lost,
  title={Lost in the middle: How language models use long contexts},
  author={Liu, Nelson F and Lin, Kevin and Hewitt, John and Paranjape, Ashwin and Bevilacqua, Michele and Petroni, Fabio and Liang, Percy},
  journal={Transactions of the association for computational linguistics},
  volume={12},
  pages={157--173},
  year={2024}
}

@article{xu2025amem,
  title={A-mem: Agentic memory for llm agents},
  author={Xu, Wujiang and Liang, Zujie and Mei, Kai and Gao, Hang and Tan, Juntao and Zhang, Yongfeng},
  journal={Advances in Neural Information Processing Systems},
  volume={38},
  pages={17577--17604},
  year={2026}
}

@inproceedings{kang-etal-2025-memory,
  title={Memory os of ai agent},
  author={Kang, Jiazheng and Ji, Mingming and Zhao, Zhe and Bai, Ting},
  booktitle={Proceedings of the 2025 Conference on Empirical Methods in Natural Language Processing},
  pages={25972--25981},
  year={2025}
}

@inproceedings{tan-etal-2025-prospect,
  title={In prospect and retrospect: Reflective memory management for long-term personalized dialogue agents},
  author={Tan, Zhen and Yan, Jun and Hsu, I-Hung and Han, Rujun and Wang, Zifeng and Le, Long and Song, Yiwen and Chen, Yanfei and Palangi, Hamid and Lee, George and others},
  booktitle={Proceedings of the 63rd Annual Meeting of the Association for Computational Linguistics (Volume 1: Long Papers)},
  pages={8416--8439},
  year={2025}
}

@inproceedings{van-etal-2026-memorai,
  title={MemORAI: Memory Organization and Retrieval via Adaptive Graph Intelligence for LLM Conversational Agents},
  author={Van, Hung Pham and Hieu, Nguyen Manh and Tuan, Khang Pham Tran and Le Hai, Nam and Van, Linh Ngo and Diep, Nguyen Thi Ngoc and Le, Trung},
  booktitle={Findings of the Association for Computational Linguistics: ACL 2026},
  pages={28235--28253},
  year={2026}
}

@inproceedings{cao-etal-2026-higmem,
  title={HiGMem: A Hierarchical and LLM-Guided Memory System for Long-Term Conversational Agents},
  author={Cao, Shuqi and He, Jingyi and Tan, Fei},
  booktitle={Findings of the Association for Computational Linguistics: ACL 2026},
  pages={33853--33862},
  year={2026}
}

@inproceedings{banerjee-etal-2026-apex,
  title={APEX-MEM: Agentic Semi-Structured Memory with Temporal Reasoning for Long-Term Conversational AI},
  author={Banerjee, Pratyay and Moshtaghi, Masud and Subramanian, Shivashankar and Misra, Amita and Chadha, Ankit},
  booktitle={Proceedings of the 64th Annual Meeting of the Association for Computational Linguistics (Volume 1: Long Papers)},
  pages={16470--16489},
  year={2026}
}

@inproceedings{zhang-etal-2026-lightweight,
  title={Lightweight llm agent memory with small language models},
  author={Zhang, Jiaquan and Zhang, Chaoning and Chen, Shuxu and Huang, Zhenzhen and Zheng, Pengcheng and Wang, Zhicheng and Guo, Ping and Mo, Fan and Bae, Sung-Ho and Zou, Jie and others},
  booktitle={Proceedings of the 64th Annual Meeting of the Association for Computational Linguistics (Volume 1: Long Papers)},
  pages={12914--12929},
  year={2026}
}

@article{liu2026simplemem,
  title={Simplemem: Efficient lifelong memory for llm agents},
  author={Liu, Jiaqi and Su, Yaofeng and Xia, Peng and Han, Siwei and Zheng, Zeyu and Xie, Cihang and Ding, Mingyu and Yao, Huaxiu},
  journal={arXiv preprint arXiv:2601.02553},
  year={2026}
}

@inproceedings{li-etal-2026-locomo,
  title={Locomo-plus: Beyond-factual cognitive memory evaluation framework for llm agents},
  author={Li, Yifei and Guo, Weidong and Zhang, Lingling and Xu, Rongman and Huang, Muye and Liu, Hui and Xu, Lijiao and Xu, Yu and Liu, Jun},
  booktitle={Proceedings of the 64th Annual Meeting of the Association for Computational Linguistics (Volume 1: Long Papers)},
  pages={25085--25100},
  year={2026}
}

@article{zhang2025qwen3embedding,
  title={Qwen3 embedding: Advancing text embedding and reranking through foundation models},
  author={Zhang, Yanzhao and Li, Mingxin and Long, Dingkun and Zhang, Xin and Lin, Huan and Yang, Baosong and Xie, Pengjun and Yang, An and Liu, Dayiheng and Lin, Junyang and others},
  journal={arXiv preprint arXiv:2506.05176},
  year={2025}
}

@misc{openai2025gpt41,
  title={Introducing GPT-4.1 in the API},
  author={OpenAI and Kumar, Ananya and Yu, Jiahui and Hallman, John and others},
  year={2025},
  url={https://openai.com/index/gpt-4-1/}
}

@misc{openai2024embeddings,
  title={New embedding models and API updates},
  author={OpenAI},
  year={2024},
  url={https://openai.com/index/new-embedding-models-and-api-updates/}
}

@article{mcnemar1947note,
  title={Note on the sampling error of the difference between correlated proportions or percentages},
  author={McNemar, Quinn},
  journal={Psychometrika},
  volume={12},
  number={2},
  pages={153--157},
  year={1947},
  publisher={Cambridge University Press \& Assessment}
}

% ============================================================
% Appendix
% ============================================================

\appendix

% ============================================================
% Implementation Details
% ============================================================

\section{Implementation Details}

This section provides additional implementation details for the main
components of \method{}.
We present condensed versions of the core prompting instructions;
formatting requirements, structured-output constraints, and
implementation-level validation instructions are omitted for clarity.

\subsection{Prompted Components}
\paragraph{Memory extraction.}
Memory is constructed independently for each conversation session.
Generative memory construction uses only user-side content, while
assistant-generated content remains accessible from the original
conversation through source recovery when needed.
For LoCoMo, which represents conversations between two participants
rather than in user--assistant format, both participants are mapped to
the user side before memory extraction, allowing the same extraction
pipeline to be applied across benchmarks.
The extractor converts the resulting conversational content into
compact, subject-explicit memories that remain interpretable outside
their original context.
Its core instruction is summarized as follows:

\begin{quote}
\small
Extract durable, answer-bearing information from the supplied
conversation session as atomic memories.
Each memory should describe one independently retrievable fact, event,
preference, state, measurement, relation, list, or interaction, with
an explicit subject.

Preserve exact entities, dates, values, durations, and relations when
they are essential to the meaning.
Resolve local references from the surrounding conversation, but do not
infer unsupported information or use outside knowledge.
Distinguish completed events from plans, ongoing states, and stable
facts.
When important interaction content cannot be represented compactly,
retain a concise memory that can be linked back to the original
conversation.
Return structured JSON only.
\end{quote}

The extractor receives the session identifier, session timestamp,
and conversational messages.
Each extracted memory contains five fields:
\texttt{subject}, \texttt{fact}, \texttt{event\_date},
\texttt{status}, and \texttt{kind}.

A representative output is:

\begin{verbatim}
{
  "<session_id>": {
    "memories": [{
      "subject": "User",
      "fact": "prefers small, quiet hotels",
      "event_date": "unknown",
      "status": "stable",
      "kind": "preference"
    }]
  }
}
\end{verbatim}

Provenance is retained separately from the compact memory content.
A card can therefore serve directly as retrieval evidence while
retaining a link to its originating conversation when higher-fidelity
evidence is required.

\paragraph{Query planning.}
The planner analyzes the evidence requirements of a query before
retrieval.
Rather than generating an answer, it identifies the entities,
temporal constraints, operation, and access requirements needed to
retrieve the relevant evidence.
Its core instruction is summarized as follows:

\begin{quote}
\small
Analyze the question in terms of the evidence required to answer it.
Identify the target entities, temporal constraints, and operation to
be performed over memory evidence.

Determine whether the required evidence is expected to span multiple
memories or sessions, and separately whether the query requires
source-level content that compact memory may not preserve.
For queries requiring broader discovery, generate short retrieval
rewrites that preserve the entities and constraints of the original
question without predicting the answer.
Return a structured retrieval plan rather than answering the question.
\end{quote}

The planner receives the current question together with its query
timestamp.
Its structured output specifies retrieval routes, operation, answer
mode, target entities, temporal scope, multi-session demand,
source-recovery demand, and optional retrieval rewrites.

The access configuration is deterministically derived from these
structured requirements.
Queries supported by local compact evidence use \lookup{};
queries requiring distributed or compositional evidence use
\compose{}; and queries requiring fidelity-sensitive source content
use \replay{}.
At most two answer-free retrieval rewrites are retained.

\paragraph{Evidence-constrained answering.}
The answer model receives the query, query time, planner-derived
operation, and evidence produced by the selected access configuration.
For \replay{}, the evidence additionally contains recovered
provenance-linked source content.
The core answering instruction is summarized as follows:

\begin{quote}
\small
Answer the question using the supplied long-term memory evidence.
Inspect all relevant evidence before answering and combine multiple
memories when required by the requested operation.

Preserve entity identity, temporal relations, and exact values when
relevant.
For aggregation and comparison, operate only over distinct evidence
matching the requested entities and constraints.
For latest or updated information, resolve conflicting evidence using
temporal order.
Do not introduce information unsupported by the supplied evidence.
Return a concise structured answer.
\end{quote}

Operation-specific instructions further guide temporal normalization,
deduplication, aggregation, comparison, and evidence integration
according to the planner output.

\subsection{Retrieval and Source Recovery}

Memory cards are represented using 1,536-dimensional
text-embedding-3-small embeddings and reranked with Qwen3-Rerank.
Retrieval combines semantic similarity with lightweight lexical,
temporal, and card-aware signals.

For \compose{}, multiple query views contribute candidates before
deduplication and reranking.
The resulting candidate pool is reranked against the original
question and reduced to the same compact evidence budget used
throughout the main experiments.
Discovery can therefore expand without increasing the final number of
compact cards presented to the answer model.

For fidelity-sensitive queries, \replay{} uses retrieved compact cards
to locate provenance-linked source conversations.
The localized source content is appended to the compact evidence for
higher-fidelity reading.
Source recovery is query-gated: recoverable provenance alone does not
trigger context expansion.

\subsection{Memory Statistics}

Table~\ref{tab:memory-statistics} summarizes the atomic-memory
representations used in the primary experiments.
Despite the length of the underlying conversation histories,
individual cards remain compact, averaging fewer than 20 tokens on
both benchmarks.

\begin{table}[h]
\centering
\small
\setlength{\tabcolsep}{3pt}
\begin{tabular}{@{}lrrrr@{}}
\toprule
Dataset & Sess. & Cards & Cards/Sess. & Tok./Card \\
\midrule
LoCoMo        & 272    & 3,804   & 13.99 & 18.01 \\
LongMemEval-S & 23,867 & 197,605 & 8.28  & 18.99 \\
\bottomrule
\end{tabular}
\caption{
Statistics of the atomic-memory representations used in the primary
experiments.
}
\label{tab:memory-statistics}
\end{table}

% ============================================================
% Replay Analysis
% ============================================================

\section{Replay Analysis}
\label{sec:replay-analysis}

We further analyze \replay{} on the 56 LongMemEval-S assistant-output
questions.
These questions provide a concentrated diagnostic for reading
fidelity because their answer-critical content is often difficult to
preserve in compact memory.
We decompose \replay{} into two stages:
localizing the answer-bearing source and answering from the recovered
content.

\begin{table}[h]
\centering
\small
\setlength{\tabcolsep}{3.2pt}
\begin{tabular}{@{}lrrr@{}}
\toprule
Condition & Acc. & Src. hit & Acc.$\mid$hit \\
\midrule
Compact cards
    & 32.14 & -- & -- \\
Query-gated \replay{}
    & \textbf{94.64} & \textbf{94.64} & \textbf{100.00} \\
\bottomrule
\end{tabular}
\caption{
Source-localization and reading analysis on the 56 LongMemEval-S
assistant-output questions.
Values are percentages.
}
\label{tab:replay-chain}
\end{table}

Compact cards alone answer 18 of the 56 assistant-output questions
correctly.
Query-gated \replay{} raises accuracy to 94.64\%, an absolute
improvement of 62.50 points.

More importantly, \replay{} successfully localizes the answer-bearing
source for 53 of the 56 questions.
All 53 questions are answered correctly once the relevant source is
recovered, yielding 100\% accuracy conditioned on source localization.
The three remaining errors correspond to failures to localize the
correct source.

This decomposition indicates that, for fidelity-sensitive
assistant-output recall, the remaining bottleneck lies in source
localization rather than in reading the recovered content.
Further improvements should therefore focus on identifying the correct
provenance-linked source rather than increasing source-context length.

\end{document}